\documentclass[10pt,twocolumn,letterpaper]{article}

\usepackage{wacv}              

\usepackage{graphicx}
\usepackage{amsmath}
\usepackage{amssymb}
\usepackage{booktabs}

\usepackage{algorithm}
\usepackage{algorithmic}
\usepackage{multirow}
\usepackage{paralist}
\usepackage{arydshln}
\usepackage{float}
\usepackage{makecell}  
\usepackage{arydshln}
\usepackage{amssymb} 

\usepackage[accsupp]{axessibility}
\definecolor{wacvblue}{rgb}{0.21,0.49,0.74}
\usepackage[pagebackref,breaklinks,colorlinks,allcolors=wacvblue]{hyperref}
\definecolor{emailcolor}{RGB}{54,125,189}

\def\wacvPaperID{28} 
\def\confName{WACV}
\def\confYear{2026}

\newcommand\norm[1]{\left\lVert#1\right\rVert}%
\newcommand{\mycomment}[1]{}
\newcommand{\OursApproach}{OCSPruner{}}

\title{One-Cycle Structured Pruning via Stability-Driven Subnetwork Search}

\mycomment{
\author{
Deepak Ghimire$^1$ \\
IT Application Research Center, \\
Korea Electronics Technology Institute \\
{\tt\small \href{mailto:deepak@keti.re.kr}{deepak@keti.re.kr}} 
\and
Dayoung Kil$^2$ \\
Department of Intelligent Semiconductors \\
Soongsil University \\
{\tt\small \href{mailto:dayoung-kil@naver.com}{dayoung-kil@naver.com}}
\and
Seonghwan Jeong$^1$ \\
IT Application Research Center, \\
Korea Electronics Technology Institute \\
{\tt\small \href{mailto:shjeong@keti.re.kr}{shjeong@keti.re.kr}}
\and
Jaesik Park$^3$ \\
Department of Computer Science \& Engineering \\
Seoul National University \\
{\tt\small \href{mailto:jaesik.park@snu.ac.kr}{jaesik.park@snu.ac.kr}}
\and
Seong-heum Kim$^2$\thanks{Corresponding author.} \\
Department of Intelligent Semiconductors \\
Soongsil University \\
{\tt\small \href{mailto:seongheum@ssu.ac.kr}{seongheum@ssu.ac.kr}}
}
}

\author{
\textbf{Deepak Ghimire$^1$ \quad Dayoung Kil$^2$ \quad Seonghwan Jeong$^1$ \quad Jaesik Park$^3$ \quad Seong-heum Kim$^2$\thanks{Corresponding author.} \vspace{2pt}}
\\
\fontsize{11}{13}\selectfont $^1$IT Application Research Center, Korea Electronics Technology Institute\\
\fontsize{11}{13}\selectfont $^2$Department of Intelligent Semiconductors, Soongsil University\\
\fontsize{11}{13}\selectfont $^3$Department of Computer Science \& Engineering, Seoul National University\\
\hspace*{0em}\scalebox{0.95}{\tt\small
\textcolor{emailcolor}{
\{\href{mailto:deepak@keti.re.kr}{deepak},
\href{mailto:shjeong@keti.re.kr}{shjeong}\}@keti.re.kr,
\href{mailto:dayoung-kil@naver.com}{dayoung-kil@naver.com},  
\href{mailto:seongheum@ssu.ac.kr}{seongheum@ssu.ac.kr},
\href{mailto:jaesik.park@snu.ac.kr}{jaesik.park@snu.ac.kr} } } 
}

\begin{document}
\maketitle

\begin{abstract}
Existing structured pruning typically involves multi-stage training procedures that often demand heavy computation. Pruning at initialization, which aims to address this limitation, reduces training costs but struggles with performance. To address these challenges, we propose an efficient one-cycle structured pruning framework that integrates pre-training, pruning, and fine-tuning into a single training cycle without compromising model performance, referred to as the `one-cycle approach’. The core idea is to search for the optimal sub-network during the early stages of network training, guided by norm-based group saliency criteria and structured sparsity regularization. We introduce a novel pruning indicator that identifies the stable pruning epoch by measuring the similarity between evolving pruning sub-networks across consecutive training epochs. Additionally, group sparsity regularization helps accelerate the pruning process, thereby speeding up overall training. Extensive experiments on the CIFAR-10/100 and ImageNet datasets using VGGNet, ResNet, and MobileNet architectures demonstrate that our method achieves state-of-the-art accuracy while being one of the most efficient pruning frameworks in terms of training time. Our code is available at \url{https://github.com/ghimiredhikura/OCSPruner}.
\end{abstract}    
\vspace{-10pt}
\section{Introduction}
\label{sec:intro}

\indent Over the past decade, deep neural networks (DNNs) have demonstrated superior performance at the expense of substantial memory usage, computing power, and energy consumption~\cite{he2016deep, deng2020model, zhai2022scaling, ghimire2022survey}. For compressing and deploying these complex models on low-capability devices, pruning has emerged as a particularly promising approach for many embedded and general-purpose computing platforms. Structured pruning~\cite{li2017pruning, wang2021neural, chen2021only, Shen_2022_CVPR, fang2023depgraph, chen2023otov2} and unstructured pruning~\cite{han2015learning, frankle2018the, ding2019global, sanh2020movement} represent the two predominant methods in pruning. Also, there has been increasing interest in middle-level pruning granularity~\cite{lin20221xn, zhang2022learning, park2023balanced}, which provides the capability to achieve fine-grained structured sparsity, combining the advantages of both unstructured fine-grained sparsity and structured coarse-grained sparsity.

\begin{figure}[tb]
\centering
\includegraphics[width=1.0\columnwidth]{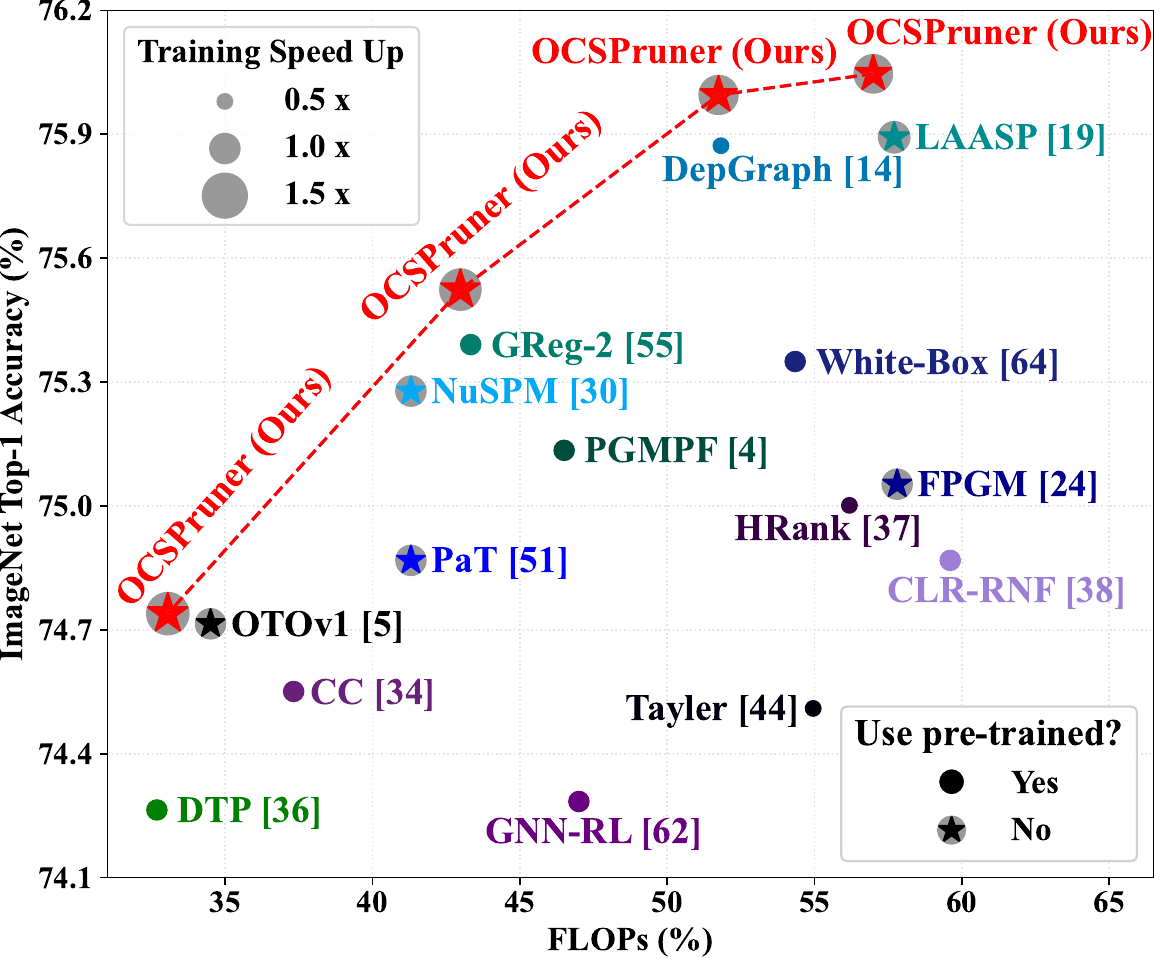}
\vspace{-15pt} 
\caption{Pruning ResNet50 on the ImageNet dataset. Each method is shown as a colored circle, with an overlaid star indicating the variant trained from scratch. The marker size reflects the training speedup relative to the baseline training cost.
}
\label{fig:imagenetcifargraph}
\end{figure}

\begin{figure*}[tb]
\centering
\includegraphics[width=1.0\linewidth]{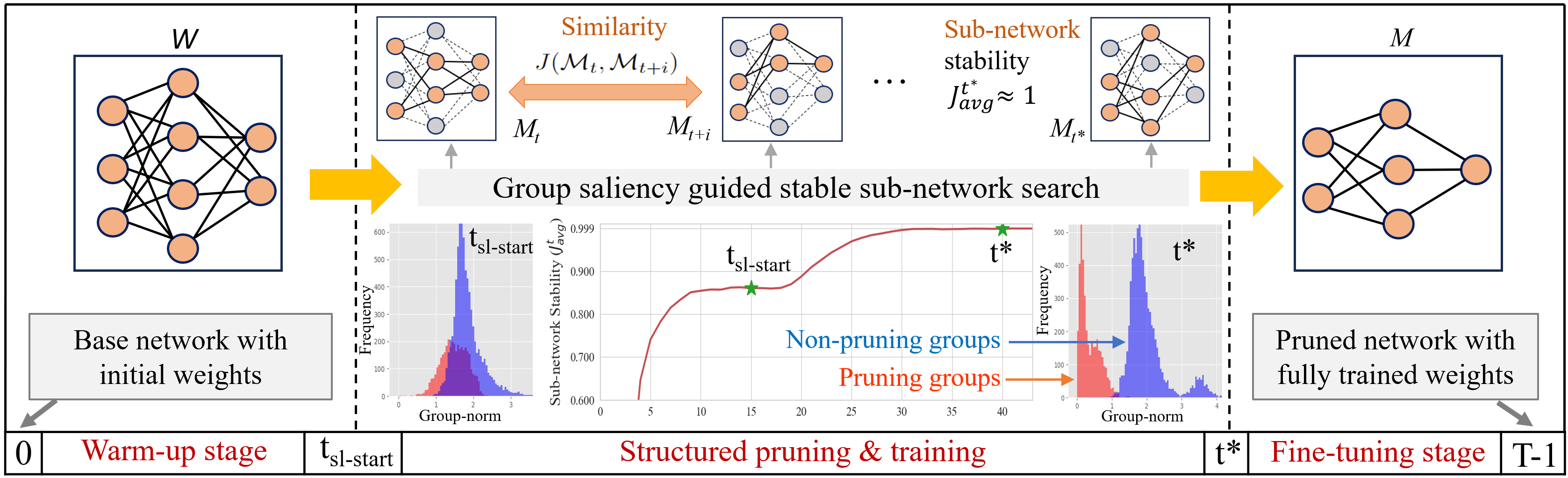}
\vspace{-15pt} 
\caption{ Overview of the \OursApproach{} algorithm. The process begins with training the baseline model from scratch. As training progresses, from $t_{\text{sl-start}}$ epoch, the pruning sub-structure gradually stabilizes as indicated by the sub-network stability score computed over consecutive training epochs. Final pruning occurs at the stable epoch $t^*$, followed by the remaining training epochs to converge the pruned structure.
}
\label{fig:algorithm_flow}
\end{figure*}

\indent With the emergence of large-scale DNN models~\cite{brown2020language, openai2023gpt4, zhai2022scaling}, there is a growing need to reassess the conventional pruning approach of pre-training, pruning, and fine-tuning, which demands excessive resources. Conventional pruning often requires multi-stage pipelines with repeated ﬁne-tuning procedures or offline re-training. Without efficient learning mechanisms, it is not only a waste of computational power and time but also less applicable to real-time scenarios. One-cycle training and pruning, also known as one-cycle pruning, presents itself as a viable alternative solution to this problem. It is therefore important to explore whether a one-cycle pruning framework can quickly adapt to new tasks or environments in real-world applications such as autonomous vehicles, drones, and robots~\cite{ral24ghafourian, chen2021only}. \\ 
\indent To address the limitations of conventional pruning, we introduce \textbf{O}ne-\textbf{C}ycle \textbf{S}tructured \textbf{Pruner} (\textbf{OCSPruner}), {which eliminates the requirement for pre-trained base network for pruning.
Considering scenarios where networks need to learn from scratch within limited resource budgets, the efficiency gained from avoiding pre-training the base model becomes more compelling.} Hence, we focus on efficient and stable structured pruning, ensuring model performance in a more sustainable way. Basically, structured pruning offers the advantage of creating slimmer yet structured networks that do not rely on specific hardware or software; however, it can occasionally lead to performance drops, particularly when dealing with high pruning ratios. \\
\indent Considering the insights from the Lottery Ticket Hypothesis~\cite{frankle2018the}, which suggests that sparse, trainable subnets can be identified within larger networks, the timing and method of pruning become critical. Our focus on one-cycle structured pruning involves pruning the network at initialization~\cite{frankle2018the, lee2018snip, jorge2021progressive}. However, this type of method may suffer from potential sub-optimal network structures that result in performance degradation. {Thus, rather than pruning at initialization, our proposed strategy involves early-phase pruning during training while still maintaining a one-cycle approach.~Despite related methods like pruning-aware training (PaT)~\cite{Shen_2022_CVPR} and loss-aware automatic selection of pruning criteria (LAASP)~\cite{GHIMIRE2023104745}, they overlook structural parameter grouping during saliency estimation, leading to reduced performance~\cite{fang2023depgraph}.~Specifically, PaT~\cite{Shen_2022_CVPR} focuses solely on total channel numbers, not considering individual channel identities during stable sub-network selection, while LAASP~\cite{GHIMIRE2023104745} lacks stability estimation criteria and involves complex iterative pruning.~This paper addresses these limitations and introduces additional algorithmic enhancements for early-phase structured pruning during training from scratch, resulting in state-of-the-art performance.} \\
\indent~\cref{fig:algorithm_flow} illustrates the overall flow of the proposed pruning algorithm. Our algorithm begins with network training from scratch over a few epochs. Following this, we applied structured sparsity regularization targeting pruning groups of parameters alongside network training. For each training epoch, we proposed a novel method to measure the stability of network pruning, leveraging the similarity among temporarily pruned sub-structures across consecutive training epochs. The sub-structures are determined by globally partitioning the structurally grouped parameters into pruning and non-pruning groups utilizing norm-based group saliency scores. As training progresses, the penalty factor for regularization progressively increases, thereby promoting the partitioned group of pruning parameters to approach zero gradually. Finally, the proposed algorithm automatically determines the epoch of stable pruning, during which final pruning is executed. This will occur when the partitioning of parameters into pruning and non-pruning groups achieves stability across successive training epochs. Following this, standard training will persist with the pruned sub-structure for the remaining epochs. This approach leads to an efficient one-cycle training framework, achieving a fully trained pruned network starting from the randomly initialized base network.

The main contributions are summarized as follows: 
\begin{compactitem} 
   \item We proposed an efficient structured pruning algorithm that combines traditional training, pruning, and fine-tuning steps into a single training cycle through the pruning while training approach. 
   \item A robust pruning paradigm is suggested, which continuously monitors the evolving sub-structure guided by a pruning stability indicator to achieve an optimal architecture during the early stages of network training.
   \item The proposed pruning algorithm is validated on VGGNet, ResNet, and MobileNet architectures. Notably, it achieves 75.49\% top-1 and 92.63\% top-5 accuracy for the ResNet50 model on the ImageNet dataset, with 1.38× faster training and 57\% fewer FLOPs.
\end{compactitem}


\section{Related Work}
\label{sec:relatedwork}
\noindent\textbf{Structured Pruning.} While unstructured pruning has been a subject of long-standing research~\cite{NIPS1989_6c9882bb, NIPS1992_303ed4c6, han2015learning}, structured pruning gained popularity with the emergence of modern DNNs~\cite{li2017pruning}. Structured pruning revolves around extracting efficient sub-structures from complete models while optimizing storage and inference processes. This can involve incorporating regularization techniques to induce sparsity~\cite{louizos2017learning, ye2018rethinking} or defining criteria to identify and remove less crucial parameter groups~\cite{li2017pruning, he2022filter, ghimire2022magnitude}, or a combination of both~\cite{he2019filter, wang2021neural, fang2023depgraph}. The practice of pruning at initialization~\cite{frankle2018the, lee2018snip, jorge2021progressive}, pruning during training~\cite{Shen_2022_CVPR, chen2021only, chen2023otov2, GHIMIRE2023104745}, and pruning the pre-trained model~\cite{ding2018auto, wang2021neural, wang2019structured, ghimire2022magnitude, ye2023performance} also represents an extensively explored area in network pruning.

\noindent\textbf{Regularization in Pruning.} Regularization is a common approach to learning sparsity in DNNs~\cite{louizos2017learning, liu2017learning, ye2018rethinking}. The introduction of a penalty factor controls the level of regularization. When a modest penalty factor is uniformly applied to all weights, it facilitates the gradual acquisition of regular sparsity~\cite{wen2016learning, fang2023depgraph}. Conversely, adopting a larger or progressively increasing penalty factor targeted at a specific parameter group yields large regularization effects~\cite{ding2018auto, wang2019structured, wang2021neural}. Moreover, while utilizing parameter saliency is a widely adopted method for direct pruning~\cite{li2017pruning, GHIMIRE2023104745}, the combination of parameter saliency and regularization is often used for the gradual learning of structural sparsity~\cite{ding2018auto, wang2019structured, wang2021neural, fang2023depgraph}. While tuning task and model-specific hyperparameters has been challenging in regularization-based pruning, the advent of automatic parameter tuning algorithms using AutoML~\cite{falkner2018bohb} and NAS~\cite{wang2023prenas} can simplify the process.

\noindent\textbf{One-cycle Pruning.} The pruning of a pre-trained network and fine-tuning the resulting pruned architecture is a commonly used framework for network slimming. However, this approach can be computationally intensive due to the multiple rounds of network training it entails. An alternative approach, known as pruning at initialization, quickly gained popularity and includes methods such as Lottery Ticket Hypothesis~\cite{frankle2018the} and its variants~\cite{bai2022dual, pmlr-v206-xiong23a}, single-shot network pruning (SNIP)~\cite{lee2018snip}, and progressive skeletonization~\cite{jorge2021progressive}. However, some of its limitations are the risk of sub-optimal network structures, potential performance gaps, and difficulty in defining pruning criteria~\cite{frankle2018the}. Another framework involving pruning and training in just a single training cycle includes methods such as PaT~\cite{Shen_2022_CVPR}, LAASP~\cite{GHIMIRE2023104745}, and only-train-once (OTO)~\cite{chen2021only, chen2023otov2}. Although these approaches are more informative and have great potential compared to pruning at initialization, there is little attention in the literature. Therefore, in this paper, we further investigate the one-cycle pruning approach and develop an efficient structured pruning algorithm while achieving high accuracy.
\section{Methods}
\label{sec:methods}

\subsection{Problem Formulation}

Let us assume a convolutional neural network (CNN) with $L$ layers, each layer parameterized by $\mathcal{W}^l \in {\mathbb{R}^{C^l_\text{out}\times{C^l_\text{in}}\times{K^l}\times{K^l}}}$, $K$ being the kernel size and $C_\text{out}$ and $C_\text{in}$ being number of output and input channels, respectively. Given a dataset $\mathcal{D}$ consisting of $N$ input-output pairs $\{ (x_i, y_i) \}_{i=1}^N$, learning the network parameters with a given pruning ratio $\alpha$ can be formulated as the following optimization problem: 
\begin{equation}
\underset{\mathcal{M}}{\arg\min} \bigl( \mathcal{L}(\mathcal{M}, \mathcal{D})\bigl),  ~~\mbox{s.t.}~~ \frac{\Psi\Bigl(f(\mathcal{M}, x_i)\Bigl) }{\Psi\Bigl(f(\mathcal{W}, x_i)\Bigl)} \geq \alpha,
\label{eqn:pruneoptfn}
\end{equation}
where $\mathcal{M} \subset \mathcal{W}$ are the parameters after pruning, $\mathcal{L}$ is the network loss, $f()$ encodes the network function, and $\Psi()$ maps the network parameters to the pruning constraints, i.e., in our case, float point operations (FLOPs) of the network. The method can easily be scaled to other constraints, such as latency or memory.

\subsection{Group Saliency Criteria}
\label{sec:group_saliency_criteria}

Conventional saliency estimation for pruning convolutional filters considers only the weight values within this filter to measure its importance. However, in this paper, we also take into account its {structurally associated coupled parameters} for estimating the saliency, which we call group saliency. The DepGraph~\cite{fang2023depgraph} algorithm is used to analyze parameter dependencies within $W$, and entire trainable parameters are partitioned into several groups $\mathcal{G} = \{g\}$. Simple norm-based criteria~\cite{li2017pruning} is utilized for group saliency estimation. Each group $g = \{w_1, w_2, ..., ~w_{|g|}\}$ has an arbitrary number of sets of parameters with different dimensionality. Given our adoption of global pruning, we propose to incorporate \textbf{\emph{local normalization}} into the calculation of group importance, thus facilitating the global ranking of filters for the pruning process. The overall saliency score of a group, as determined by the $l_2$-norm used in this study, is defined as:
\begin{equation}
S(g) = \frac { \sum_{w \in g} \norm{w}_2 / \sqrt{|w|}} {|g|},
\label{eqn:gss}
\end{equation}
where $|.|$ denotes set cardinality. 

In~\cref{eqn:gss}, we calculate the normalized $l_2$-norm~\cite{NIPS2016_6e7d2da6} of the set of elements within a group. The resulting local norms are summed up and again normalized with the cardinality of the group. Such normalization is induced because the group size and dimensionality of the set of parameters within the groups could be different. This will ensure fair comparison within different groups or layers during global pruning. 

\subsection{Pruning Stability}

During the pruning sub-network search process, at the end of each training epoch, we temporarily prune the network for a given pruning constraint and ratio using group saliency criteria defined in~\cref{eqn:gss}. Suppose $\mathcal{M}_{t-i}$ and $\mathcal{M}_t$ denotes the pruned sub-network structures at training epoch $t-i$ and $t$, respectively. %
Those sub-networks are used to check for pruning stability defined by their similarity.
As soon as the sub-network stabilizes in subsequent training epochs, the pruning is made permanent, and the resulting pruned sub-network is further trained for convergence. 

PaT~\cite{Shen_2022_CVPR} uses a similar approach, but they check for pruning stability calculated using only the number of channels in each layer. The limitation of their approach is that the two sub-networks could be identical in terms of the number of channels in each layer. However, the identity of those channels derived from the original network could still differ, leading to unstable pruning. 
To solve this issue, in this paper, we propose assessing pruning stability using the total number of retained filters and \textbf{\emph{the identity of derived filters}} from the original network.

Suppose $\mathcal{F}_{t-i}^l$ and $\mathcal{F}_t^l$ be the set of filters in temporarily pruned sub-networks at training epoch $t-i$ and $t$, respectively. The similarity between these two sub-networks based on the cardinality of the filter set, as well as their corresponding identity, in each layer, is defined using Jaccard Similarity $J$: 
\begin{equation}
J(\mathcal{M}_{t-i}, \mathcal{M}_t) = \frac{1}{L} \sum_{l=1}^L \frac {|\mathcal{F}_{t-i}^l \cap \mathcal{F}_t^l|} {|\mathcal{F}_{t-i}^l \cup \mathcal{F}_t^l|}.
\label{eqn:jaccard}
\end{equation}

The similarity value $J$ ranges from $0$ to $1$, where $1$ means two sub-networks are the same and vice-versa. In practice, we use the average of past $r$ similarities:
\begin{equation}
J_\text{avg}^t = \frac{1}{r} \sum_{k=0}^{r-1} J(\mathcal{M}_{t-k-i}, \mathcal{M}_{t-k}).
\label{eqn:avgjaccard}
\end{equation}

\begin{figure}[t]
\centering
\includegraphics[width=1.0\columnwidth]{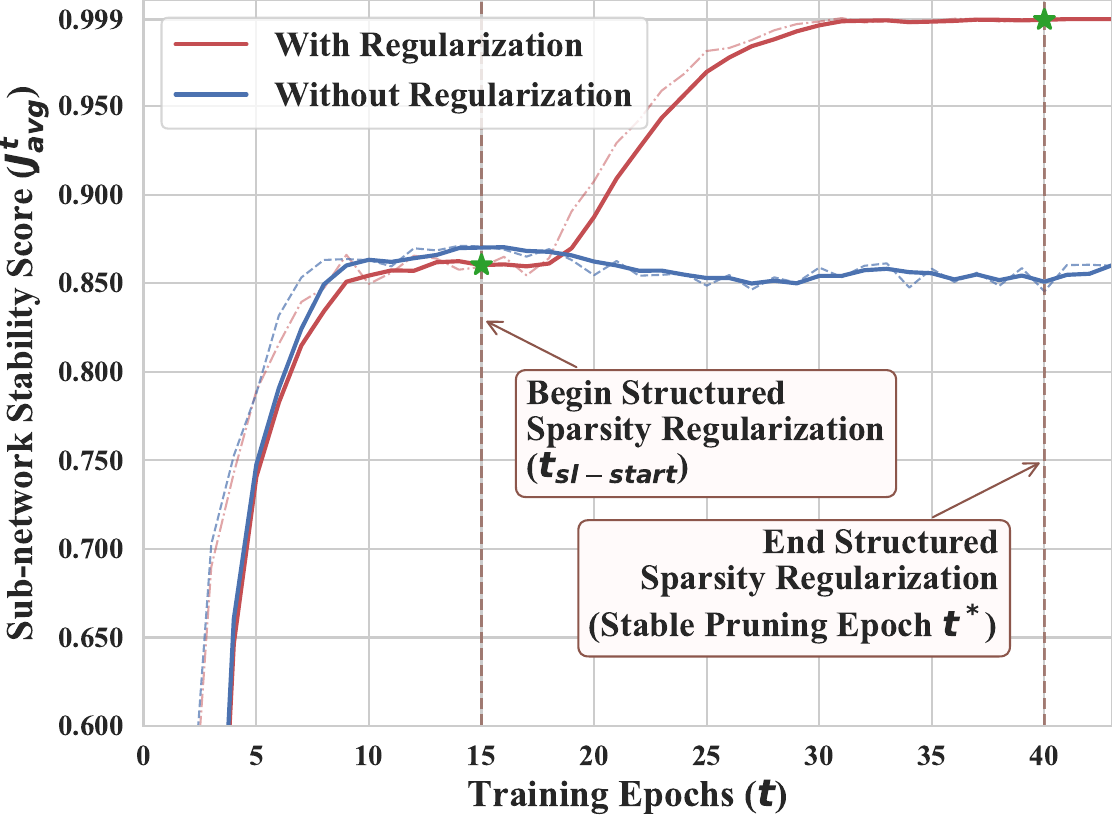}
\vspace{-15pt} 
\caption{Sub-network stability score determined with Jaccard Similarity between sub-networks acquired during training epochs. The dotted curves are the plot of raw data before averaging.}
\label{fig:fig1}
\end{figure}

The stability score determines when sparsity learning begins ($t_{\text{sl-start}}$) and is also used to identify the stable pruning epoch ($t^*$). $t_{\text{sl-start}}$ is found when consecutive epochs show minimal change in average similarity ($J_{\text{avg}}$) within a window ($r$), using a threshold value ($\tau$). Therefore, the $t_{\text{sl-start}}$ is estimated using the following expression: 
\begin{equation}
t_{\text{sl-start}} = \min\left\{ t : \left(J_{\text{avg}}^t - J_{\text{avg}}^{t-r}\right) \leq \tau \right\}.
\label{eqn:sl_start}
\end{equation}

Additionally, the stable pruning epoch ($t^*$) is estimated using the following expression:
\begin{equation}
t^* = \underset{t}{\arg\max} \Bigl( J_\text{avg}^t \geq 1-\epsilon \Bigl).
\label{eqn:spe}
\end{equation}

Here, $t^*$ is determined as the epoch where the average similarity reaches a value greater than or equal to $(1 - \epsilon)$, with $\epsilon$ representing the threshold value.

\cref{fig:fig1} shows the curves illustrating the behavior of the sub-network stability score $J_\text{avg}^t$ with and without the application of structured sparsity regularization. The figure highlights how regularization speeds up stability convergence, allowing sufficient time for fine-tuning the pruned architecture. The next section will explain this regularization technique in detail.

\subsection{Structured Sparsity Regularization}
\label{sec:sparsity_reg}

The structured sparsity regularization scheme in our approach is motivated by pruning methods DepGraph~\cite{fang2023depgraph} and GReg~\cite{wang2021neural}. DepGraph~\cite{fang2023depgraph} pruning scheme gradually sparsifies the structurally coupled parameters at the group level based on the normalized group importance scores. During the training process, the regularization is applied to all the parameter groups while driving the groups with low importance scores toward zero for pruning. On the other hand, GReg~\cite{wang2021neural} imposes a penalty term on those filters selected for pruning during the training process while gradually increasing the regularization penalty. However, the GReg~\cite{wang2021neural} technique ignores the structurally coupled parameters while imposing growing regularization for sparsity learning. 


In this study, we also employ growing regularization to enhance sparsity learning, addressing structurally linked parameters at the group level. \cref{alg:algorithm1} outlines the step-by-step procedure of the proposed pruning algorithm. In each training epoch, given group saliency scores and pruning ratio, global pruning is used to \emph{temporarily partition} the structurally grouped parameters into \emph{pruning and non-pruning groups} (Line~\ref{lst:line:partition} $\sim$~\ref{lst:line:tempprune}). The \emph{binary search} algorithm is applied to quickly partition the groups for a given pruning ratio. In every training epoch, if a group's saliency score, which was previously partitioned as a pruning group, increases, it may move to the non-pruning group, and vice versa. Then, our algorithm applies \emph{regularization to the newly partitioned pruning groups}, leaving the non-pruning groups unaffected (Line~\ref{lst:line:regularization}). 

\begin{algorithm}[t]
\caption{Algorithm of \OursApproach{}}
\small 
\begin{algorithmic}[1] \label{alg:algorithm1}
\renewcommand{\algorithmicrequire}{\textbf{Input:}}
\renewcommand{\algorithmicensure}{\textbf{Output:}}
\REQUIRE Training data $\mathcal{D}$, base model with randomly initialized weights $\mathcal{W}$, pruning ratio $\alpha$, total epochs $T$, window size for averaging stability scores $r$, sparsity learning start epoch threshold $\tau$, pruning stability threshold $\epsilon$
\ENSURE Pruned model with fully trained weights $\mathcal{M}$
\STATE $\text{SL\_STAGE} \gets$ False
\FOR {$t = 0$, $\dots$, $T-1$}
    \STATE Partition model $\mathcal{W}$ into structural groups $\mathcal{G}$ \label{lst:line:partition}
    \STATE Get group saliency scores $S(g)$ using ~\cref{eqn:gss}
    \STATE Obtain the pruned model $\mathcal{M}_t$ and pruning groups $\mathcal{G}_{\text{prune}}^t$ at pruning ratio $\alpha$. \label{lst:line:tempprune}
    \STATE Get $J_\text{avg}^t$ using ~\cref{eqn:avgjaccard}

    \IF {not $\text{SL\_STAGE}$}
        \STATE Train model $\mathcal{W}$ by gradient descent
        \IF{ $\left(J_{\text{avg}}^t - J_{\text{avg}}^{t-r}\right) \leq \tau$ }
            \STATE $\text{SL\_STAGE} \gets$ True
        \ENDIF
    \ELSIF { $ J_\text{avg}^t \geq 1-\epsilon $ }
        \STATE Set epoch $t$ as stable pruning epoch, $t^* \gets t$
        \STATE $\mathcal{M} \gets \mathcal{M}_{t^*}$
        \STATE \textbf{break}
    \ELSE
        \STATE Update regularization penalty factor (${\lambda}_t$) using~\cref{eqn:growingpenaltyfact}\label{lst:line:penalty}
        \STATE Induce structural sparsity regularization using ~\cref{eqn:l2penalty} and ~\cref{eqn:directmultiply} on $\mathcal{W}$ \label{lst:line:regularization}
        \STATE Train model $\mathcal{W}$ by gradient descent 
    \ENDIF
\ENDFOR
\FOR {$t = t^*$, $\dots$, $T-1$}
    \STATE Train pruned model $\mathcal{M}$ by gradient descent
\ENDFOR
\end{algorithmic} 
\end{algorithm}

In early network training, weights are volatile, thus, large regularization is not advisable. Therefore, initially, a small penalty is used for sparsity learning. As training advances the penalty is also increased to speed up regularization. The growing penalty factor used in our approach for structured sparsity regularization is defined as:
\begin{equation}
{\lambda}_t = {\lambda}_{t-1} + \delta \times \lfloor \frac{t-t_{\text{sl-start}}}{\Delta t} \rfloor.
\label{eqn:growingpenaltyfact}
\end{equation}

In~\cref{eqn:growingpenaltyfact}, $t$ represents the training epoch, and $\delta$ denotes the growing factor. The term $t_{\text{sl-start}}$ corresponds to the starting epoch for structured sparsity regularization. From $t_{\text{sl-start}}$ onward, the penalty factor increases by $\delta$ in every $\Delta t$ epoch interval, as determined by the floor function $\lfloor~\rfloor$ (Line~\ref{lst:line:penalty}). 


Suppose that, at training epoch $t$, $\mathcal{G}_{\text{prune}}^t$ and $\mathcal{G}_{\text{non-prune}}^t$ denote the partitioning of structural groups into pruning and non-pruning groups. First, similar to GReg~\cite{wang2021neural}, we impose a growing $l_2$-norm penalty estimated using a pruning group of parameters to the network's original loss function defined as:
\begin{equation}
{\mathcal{L}}_{\text{total}}^t = {\mathcal{L}}_{\text{orig}}^t + \lambda_t  \sum_{g \in \mathcal{G}_{\text{prune}}^t} \sum_{w \in g} \norm{w}_2
\label{eqn:l2penalty},
\end{equation}
where ${\mathcal{L}}_{\text{orig}}^t$ is the network original loss before imposing penalty term. 

The structured sparsity regularization imposed by~\cref{eqn:l2penalty} gradually drives the pruning group parameters to zero. But, as our algorithm intends to finalize the pruning process as early as possible, we define an additional sparsity learning scheme to further drive the pruning group of parameters to zero. To achieve this, we propose to directly multiply pruning group parameters with a multiplication factor estimated using the network's current learning rate and growing penalty factor as follows:
\begin{equation}
\mathcal{G}_{\text{prune}}^t = \{g (1 - {\lambda}_t \times \text{{learning-rate}}_t) \mid g \in \mathcal{G}_{\text{prune}}^t\}.
\label{eqn:directmultiply}
\end{equation}


In the early stage, sparsity learning,~\cref{eqn:directmultiply} gently reduces weights, then gradually intensifies. This two-step process provides smooth yet fast regularization, making pruning stable early while leaving enough time for fine-tuning the pruned model.
\section{Results and Discussion}
\label{sec:result}
\subsection{Experimental Setup}

To evaluate our pruning algorithm, we conducted experiments on the CIFAR-10/100~\cite{krizhevsky2009learning} and ImageNet~\cite{russakovsky2015imagenet} datasets. Initially, we pruned the relatively simple single-branch VGGNet~\cite{Simonyan15} on the CIFAR-10 and CIFAR-100 datasets. Next, we evaluate the pruning of more complex multi-branch ResNet~\cite{he2016deep} models, on both CIFAR-10 and ImageNet. 
Finally, we evaluated the 
pruning of a compact network, MobileNetV2~\cite{sandler2018mobilenetv2}, on the ImageNet.

Both CIFAR and ImageNet datasets are trained with a Stochastic Gradient Descend (SGD) optimizer with a momentum of 0.9 using data argumentation strategies adapted from official PyTorch~\cite{paszke2017automatic} examples. For training on CIFAR-10/100 datasets, the models are trained for 300 epochs with batch size 128, in which the learning rate follows the MultiStepLR scheduler. Conversely, for training on the ImageNet dataset, the learning rate follows a cosine function-shaped decay strategy, gradually approaching zero. The training settings for MobileNetV2 on the ImageNet dataset are adapted from HBONet~\cite{li2019hbonet}. Please refer to the supplement for more comprehensive information on the training and pruning parameters.

For the CIFAR dataset, we observe slight variations across training trials; therefore, we report the average over three runs. In contrast, ImageNet results were consistent across runs; hence, results from a single trial are reported.

\subsection{VGG16/19 on CIFAR-10/100}
\cref{tab:vggprune} compares the pruning of VGG16/19 on CIFAR-10/100 datasets using our method against several state-of-the-art approaches.~The VGG16 pruning on CIFAR-10 is compared with RCP~\cite{li2022revisiting}, LAASP~\cite{GHIMIRE2023104745}, PGMPF~\cite{Cai_An_Yang_Yan_Xu_2022}, CPGCN~\cite{ijcai2022p431}, OTO~\cite{chen2021only,chen2023otov2}, DLRFC~\cite{he2022filtereccv}, and DCFF~\cite{lin2023training}. Among these methods, similar to ours, OTO, and DCFF also pruned the network from scratch. While OTO, CPGCN, and DLRFC achieve higher parameter reduction rates, their top-1 accuracy is lower than ours. Notably, OTO uses a specially designed training optimizer, whereas our method works with the standard SGD optimizer. With only $26.01\%$ and $21.22\%$ of the original network FLOPs remaining, we achieved top-1 accuracy of $93.88\%$ and $93.76\%$, respectively, the highest reported for similar pruning rates. Notably, while DLRFC and DCFF show improvements in pruned model performance, this is largely due to their comparison against relatively lower baseline accuracy.

\cref{tab:vggprune} also shows our method's VGG19 pruning performance in comparison with EigenDamage~\cite{wang2019eigendamage}, GReg~\cite{wang2021neural}, and DepGraph~\cite{fang2023depgraph} on the CIFAR-100 dataset. We achieved the highest top-1 accuracy of $70.47\%$ while reducing the network parameters around $95\%$ and FLOPs around $89\%$. Although GReg~\cite{wang2021neural} also utilizes the concept of growing regularization and uses a fully trained network for pruning, its top-1 accuracy is nearly 3\% lower than ours. 

\begin{table}[tb]
\caption{Pruning of VGG16/19 on CIFAR-10/100 dataset. Our results are the average outcome from three trials.}
\label{tab:vggprune}
\vspace{-5pt} 
\centering
\small 
\setlength{\tabcolsep}{1.6pt} 
\begin{tabular}{@{}llp{0.7cm}lllrr@{}}
\toprule
\begin{tabular}[c]{@{}l@{}}Model/\\Dataset\end{tabular} & Method & \begin{tabular}[l]{@{}l@{}}Pre-\\train?\end{tabular} & \begin{tabular}[c]{@{}c@{}}FLOPs\\(\%)\end{tabular} & \begin{tabular}[c]{@{}c@{}}Params\\(\%)\end{tabular} & \begin{tabular}[c]{@{}c@{}}Base\\Acc.\\(\%)\end{tabular} & \begin{tabular}[c]{@{}c@{}}Pruned\\Acc.\\(\%)\end{tabular} & \begin{tabular}[c]{@{}c@{}}Acc.\\Drop\\(\%)\end{tabular} \\
\midrule
\multirow{10}{*}{\rotatebox[origin=c]{90}{\shortstack{VGG16/\\CIFAR-10}}} & RCP~\cite{li2022revisiting} & \checkmark & 47.99 & 42.11 & 94.33 & {93.94} & 0.39 \\
& PGMPF~\cite{Cai_An_Yang_Yan_Xu_2022} & \checkmark & 34.00 & - & 93.68 & 93.60 & 0.08\\
& CPGCN~\cite{ijcai2022p431} & \checkmark & 26.93 & {6.94} & 93.10 & 93.08 & 0.02\\
& LAASP~\cite{GHIMIRE2023104745} & $\times$ & {39.54} & {26.83} & 93.79 & {93.79} & 0.00\\
& OTOv1~\cite{chen2021only} & $\times$ & 26.80 & {5.50} & 93.20 & 93.30 & -0.10\\
& {\textbf{\OursApproach{}}} & $\times$ & {26.01} & 9.72 & 94.08 & {93.88} & 0.20 \\
\cdashline{2-8}
& DLRFC~\cite{he2022filtereccv} & \checkmark & 23.05 & 5.62 & 93.25 & {93.64} & -0.39\\
& OTOv2~\cite{chen2023otov2} & $\times$ & 23.70 & {4.90} & 93.20 & 93.20 & 0.00\\
& DCFF~\cite{lin2023training} & $\times$ & 23.13 & 7.20 & 93.02 & 93.47 & -0.45\\
& {$\textbf{\OursApproach{}}$} & $\times$ & {21.22} & 7.46 & 94.08 & {93.76} & 0.32 \\
\midrule
\multirow{4}{*}{\rotatebox[origin=c]{90}{\shortstack{VGG19/\\CIFAR-100}}} & ED~\cite{wang2019eigendamage} & \checkmark & 11.37 & - & 73.35 & 65.18 & 8.17\\
& GReg-2~\cite{wang2021neural} & \checkmark & 11.31 & - & 74.02 & 67.75 & 6.27\\
& DepGraph~\cite{fang2023depgraph} & \checkmark & {11.21} & - & 73.50 & 70.39 & 3.11\\
& {$\textbf{\OursApproach{}}$} & $\times$ & 11.23 & {5.49} & 74.24 & 70.47 & 3.77 \\
\bottomrule
\end{tabular}
\end{table}

\subsection{ResNet on CIFAR-10}

The pruning of ResNet on CIFAR-10 using our method consistently outperforms state-of-the-art methods across depths 56 and 110 (\cref{tab:resnetcifar10}). 
While our proposed algorithm automatically learns slimmer sub-networks from scratch, we not only compare its performance with other automatic methods like FPGM~\cite{he2019filter}, and LAASP~\cite{GHIMIRE2023104745} but also assess its effectiveness against various other state-of-the-art techniques such as 
DLRFC~\cite{he2022filtereccv}, HRank~\cite{lin2020hrank}, ResRep~\cite{ding2021resrep}, ATO~\cite{Wu_2024_CVPR}, DCFF~\cite{lin2023training}, DepGraph~\cite{fang2023depgraph}, GReg~\cite{wang2021neural}, C-SGD~\cite{ding2019centripetal}, and Rethink~\cite{liu2018rethinking}.
Similar to ours, LAASP~\cite{GHIMIRE2023104745} employs a partially trained network for pruning. However, it ignores structurally coupled parameters while estimating filter saliency scores and adapts manually defined stable pruning epochs. In contrast to our technique, which gradually moves pruning parameters towards zero, FPGM~\cite{he2019filter} directly sets filters with low saliency scores to zero, while still allowing updates in the next training epoch. We prune 
ResNet56 and ResNet110 models for two different FLOPs reduction rates. 
Notably, with $38.88\%$ reduced FLOPs, ResNet56 achieves $93.65\%$ top-1 accuracy. Again, ResNet110 achieves a remarkable $94.13\%$ top-1 accuracy with only 33.10\% FLOPs.

\begin{table}[t]
\caption{Pruning ResNets on CIFAR-10 dataset. Our results are the average outcome from three trials.}
\label{tab:resnetcifar10}
\vspace{-5pt} 
\centering
\small 
\setlength{\tabcolsep}{1.6pt} 
\begin{tabular}{llp{0.7cm}lllrr}
\toprule
Depth & \begin{tabular}[r]{@{}l@{}}Method\end{tabular} & \begin{tabular}[c]{@{}l@{}}Pre-\\train?\end{tabular} & \begin{tabular}[c]{@{}c@{}}FLOPs\\(\%)\end{tabular} & \begin{tabular}[c]{@{}c@{}}Params\\(\%)\end{tabular} & \begin{tabular}[c]{@{}c@{}}Base\\Acc.\\(\%)\end{tabular} & \begin{tabular}[c]{@{}c@{}}Pruned\\Acc.\\(\%)\end{tabular} & \begin{tabular}[c]{@{}c@{}}Acc.\\Drop\\(\%)\end{tabular} \\
\midrule
\multirow{13}{*}{56} & HRank~\cite{lin2020hrank} & \checkmark & 50.00 & - & 93.26 & 93.17 & 0.09\\
& DLRFC~\cite{he2022filtereccv} & \checkmark & 47.42 & {44.37} & 93.06 & 93.57 & -0.51\\
& ResRep~\cite{ding2021resrep} & \checkmark & {47.09} & - & 93.71 & 93.71 & 0.00 \\
& NuSPM~\cite{Lee_2024_WACV} & $\times$ & 49.72 & - & - & 93.50 & - \\
& FPGM~\cite{he2019filter} & $\times$ & 47.40 & - & 93.59 & 92.93 & 0.66\\
& LAASP~\cite{GHIMIRE2023104745} & $\times$ & 47.40 & 61.43 & 93.61 & 93.45 & 0.16 \\
& {$\textbf{\OursApproach{}}$} & $\times$ & {46.93} & {50.27} & 93.97 & {93.80} & 0.17\\
\cdashline{2-8}
& GReg-2~\cite{wang2021neural} & \checkmark & 39.21 & - & 93.36 & 93.36 & 0.00 \\
& C-SGD~\cite{ding2019centripetal} & \checkmark & 39.15 & - & 93.39 & 93.44 & -0.05\\
& DepGraph~\cite{fang2023depgraph} & \checkmark & 38.91 & - & 93.53 & 93.64 & -0.11\\
& ATO~\cite{Wu_2024_CVPR} & $\times$ & 45.00 & - & 93.50 & 93.74 & -0.24\\
& DCFF~\cite{lin2023training} & $\times$ & 43.75 & 44.71 & 93.26 & 93.26 & 0.00 \\
& {$\textbf{\OursApproach{}}$} & $\times$ & {38.88} & {41.42} & 93.97 & {93.65} & 0.32\\
\midrule
\multirow{6}{*}{110} & FPGM~\cite{he2019filter} & $\times$ & 47.70 & - & 93.68 & 93.74 & -0.06\\
& {$\textbf{\OursApproach{}}$} & $\times$ & {46.37} & {52.71} & 94.36 & {94.30} & 0.06\\
\cdashline{2-8}
& HRank~\cite{lin2020hrank} & \checkmark & 41.80 & - & 93.50 & 93.36 & 0.14 \\
& LAASP~\cite{GHIMIRE2023104745} & $\times$ & 41.29 & 54.56 & 94.41 & 93.61 & 0.80 \\
& DCFF~\cite{lin2023training} & $\times$ & {33.18} & {32.37} & 93.50 & {93.80} & -0.30\\
& {$\textbf{\OursApproach{}}$} & $\times$ & {33.10} & {34.93} & 94.36 & {94.13} & 0.23 \\
\bottomrule
\end{tabular}
\end{table}

\begin{table*}[t]
\caption{Pruning ResNet50/MobileNetV2 on ImageNet dataset.}
\label{tab:resnet50-mobilenetv2-imagenet}
\vspace{-8pt} 
\centering
\small 
\setlength{\tabcolsep}{2.5pt} 
\begin{tabular}{clcccccccccr}
\toprule
Model & Method & \begin{tabular}[l]{@{}l@{}}Pre-train?\end{tabular} & \begin{tabular}[c]{@{}c@{}}FLOPs (\%)\end{tabular} & \begin{tabular}[c]{@{}c@{}}Params (\%)\end{tabular} & \begin{tabular}[c]{@{}c@{}}Baseline \\Top-1 \\Acc. (\%)\end{tabular} & \begin{tabular}[c]{@{}c@{}}Pruned \\Top-1 \\Acc. (\%)\end{tabular} & \begin{tabular}[c]{@{}c@{}}Top-1 \\Acc. \\Drop (\%)\end{tabular} & \begin{tabular}[c]{@{}c@{}}Baseline \\Top-5 \\Acc. (\%)\end{tabular} & \begin{tabular}[c]{@{}c@{}}Pruned \\Top-5 \\Acc. (\%)\end{tabular} & \begin{tabular}[c]{@{}c@{}}Top-5 \\Acc. \\Drop (\%)\end{tabular} & \begin{tabular}[c]{@{}c@{}}Training \\SpeedUp\end{tabular} \\ 
\midrule
\multirow{19}{*}{\rotatebox[origin=c]{90}{ResNet50}} & CLR-RNF~\cite{lin2022pruning} & \checkmark & 59.60 & 66.20 & 76.01 & 74.85 & 1.16 & 92.96 & 92.31 & 0.65 & $\leq 0.66 \times$ \\
& FPGM~\cite{he2019filter} & $\times$ & 57.80 & - & 76.15 & 75.03 & 1.12 & 92.87 & 92.40 & 0.47 & $\leq 1.00 \times$ \\
& LAASP~\cite{GHIMIRE2023104745} & $\times$ & 57.67 & 77.29 & 76.48 & 75.85 & 0.63 & 93.14 & 92.81 & 0.33 & $1.04 \times$ \\
& {$\textbf{\OursApproach{}}_{43\%}$} & $\times$ & {56.99} & {64.45} & 76.29 & {76.00} & 0.29 & 93.04 & {92.90} & 0.14 & ${1.28 \times}$ \\
\cdashline{2-12}
& HRank~\cite{lin2020hrank} & \checkmark & 56.18 & - & 76.15 & 74.98 & 1.17 & 92.87 & 92.33 & 0.54 & $<0.50 \times$ \\
& Tayler~\cite{molchanov2019importance} & \checkmark & 54.95 & - & 76.18 & 74.50 & 1.68 & - & - & - & $< 0.50 \times$ \\
& White-Box~\cite{zhang2022carrying} & \checkmark & 54.34 & 68.63 & 76.15 & 75.32 & 0.83 & 92.96 & 92.43 & 0.53 & $\leq 0.66 \times$ \\
& DepGraph~\cite{fang2023depgraph} & \checkmark & 51.82 & - & 76.15 & 75.83 & 0.32 & - & - & - & $< 0.50 \times$ \\
& {$\textbf{\OursApproach{}}_{48\%}$} & $\times$ & {51.74} & {62.66} & 76.29 & {75.95} & 0.34 & 93.04 & {92.75} & 0.29 & ${1.30 \times}$ \\
\cdashline{2-12}
& GNN-RL~\cite{yu2022topology} & \checkmark & 47.00 & - & 76.10 & 74.28 & 1.82 & - & - & - & $\leq 0.66 \times$ \\
& PGMPF~\cite{Cai_An_Yang_Yan_Xu_2022} & \checkmark & 46.50 & - & 76.15 & 75.11 & 0.90 & 92.93 & 92.41 & 0.52 & $\leq 0.66 \times$ \\
& GReg-2~\cite{wang2021neural} & \checkmark & {43.33} & 62.30 & 76.13 & 75.36 & 0.77 & 92.86 & 92.41 & 0.45 & $\leq 0.66 \times$ \\
& PaT~\cite{Shen_2022_CVPR} & $\times$ & {41.30} & - & - & 74.85 & - & - & - & - & $> 1.00 \times$ \\
& NuSPM~\cite{Lee_2024_WACV} & $\times$ & {41.30} & - & - & 75.30 & - & - & - & - & $> 1.00 \times$ \\
& {$\textbf{\OursApproach{}}_{57\%}$} & $\times$ & {42.98} & {51.83} & 76.29 & {75.49} & 0.80 & 93.04 & {92.63} & 0.41 & ${1.38 \times}$ \\
\cdashline{2-12}
& CC~\cite{li2021towards} & \checkmark & 37.32 & 41.39 & 76.15 & 74.54 & 1.61 & 92.87 & 92.25 & 0.62 & $\leq 0.66 \times$ \\
& {DTP~\cite{li2023differentiable}} & \checkmark & {32.68} & - & 76.13 & 74.26 & 1.87 & - & - & - & ${0.66 \times}$ \\
& OTOv1~\cite{chen2021only} & $\times$ & 34.50 & {35.50} & 76.10 & {74.70} & 1.40 & 92.90 & {92.10} & 0.80 & $< 1.00 \times$ \\
& {$\textbf{\OursApproach{}}_{66\%}$} & $\times$ & {33.05} & {37.13} & 76.29 & 74.72 & 1.57 & 93.04 & 92.13 & 0.91 & ${1.41 \times}$ \\
\midrule
\multirow{7}{*}{\rotatebox[origin=c]{90}{MobileNetV2}} 
& CC~\cite{li2021towards} & \checkmark & 71.67 & - & 71.88 & 70.91 & 0.97 & - & - & - & $\leq 0.66 \times$ \\
& AMC~\cite{he2018amc} & $\times$ & {70.00} & - & 71.80 & 70.80 & 1.00 & - & - & - & - \\
& {$\textbf{\OursApproach{}}_{30\%}$} & $\times$ & 70.32 & {87.88} & 71.79 & {70.96} & 0.83 & 90.47 & {89.84} & 0.63 & ${1.20 \times}$ \\
\cdashline{2-12}
& DepGraph~\cite{fang2023depgraph} & \checkmark & {45.45} & - & 71.87 & {68.46} & 3.41 & - & - & - & $< 0.50 \times$ \\
& DCFF~\cite{lin2023training} & $\times$ & 46.67 & 74.86 & 72.00 & {68.60} & 3.40 & 90.12 & 88.13 & 1.99 & $> 1.00 \times$ \\
& LAASP~\cite{GHIMIRE2023104745} & $\times$ & {45.50} & {64.09} & 71.79 & 68.45 & 3.34 & 90.47 & {88.40} & 2.07 & $1.08 \times$ \\
& {$\textbf{\OursApproach{}}_{55\%}$} & $\times$ & {45.40} & 67.76 & 71.79 & {68.68} & 3.11 & 90.47 & {88.16} & 2.31 & ${1.20 \times}$ \\
\bottomrule
\end{tabular}
\end{table*}

\subsection{ResNet50/MobileNetV2 on ImageNet}

We evaluate the performance of ResNet50 and MobileNetV2 on the ImageNet dataset across different pruning ratios, comparing our results with those of various state-of-the-art methods from the literature (\cref{tab:resnet50-mobilenetv2-imagenet}). Our method consistently surpasses the performance of several state-of-the-art approaches. Notably, similar to our approach, FPGM~\cite{he2019filter}, PaT~\cite{Shen_2022_CVPR}, NuSPM~\cite{Lee_2024_WACV}, and OTOv1~\cite{chen2021only} also achieve a fully trained pruned network using one-cycle training from scratch. Remarkably, with a mere $43\%$ of the original FLOPs remaining in the network, our algorithm achieves the highest accuracy ($75.49\%$ top-1 and $92.63\%$ top-5) among state-of-the-art methods in the literature for similar pruning ratios. \cref{fig:imagenetcifargraph} graphically illustrates the relationship between network FLOPs and top-1 accuracy, providing an approximation of the training speed-up associated with each pruning method in reference to the baseline network training. This clearly shows that our proposed technique consistently achieves higher top-1 accuracy compared to several other state-of-the-art methods while also being the most efficient pruning framework among them.

\cref{tab:resnet50-mobilenetv2-imagenet} also presents the pruning outcomes for MobileNetV2 on ImageNet and compares them with several state-of-the-art methods, including CC~\cite{li2021towards}, AMC~\cite{he2018amc}, 
DCFF~\cite{lin2023training}, LAASP~\cite{GHIMIRE2023104745}, and DepGraph~\cite{fang2023depgraph}. We report results using the standard $1.0\times$ MobileNetV2 for fair comparison. Experiments are conducted at $30\%$ and $55\%$ pruning ratios to align with commonly used benchmarks. In both instances, our proposed method attains superior top-1 accuracy compared to the other methods. 

The last column in \cref{tab:resnet50-mobilenetv2-imagenet} presents the comparison results for training speed enhancement (higher the better) relative to the baseline ResNet50 model training cost. As shown, our proposed technique achieved up to a ${1.41 \times}$ increase in training speed while pruning 67\% of ResNet50 FLOPs. We approximated training speedup for state-of-the-art methods based on factors such as the total number of training rounds and whether pruning was done iteratively or in one shot. This comparative speedup analysis demonstrates that our method is the most efficient among the existing techniques.

\begin{figure}[t]
\centering
\includegraphics[width=1.0\linewidth]{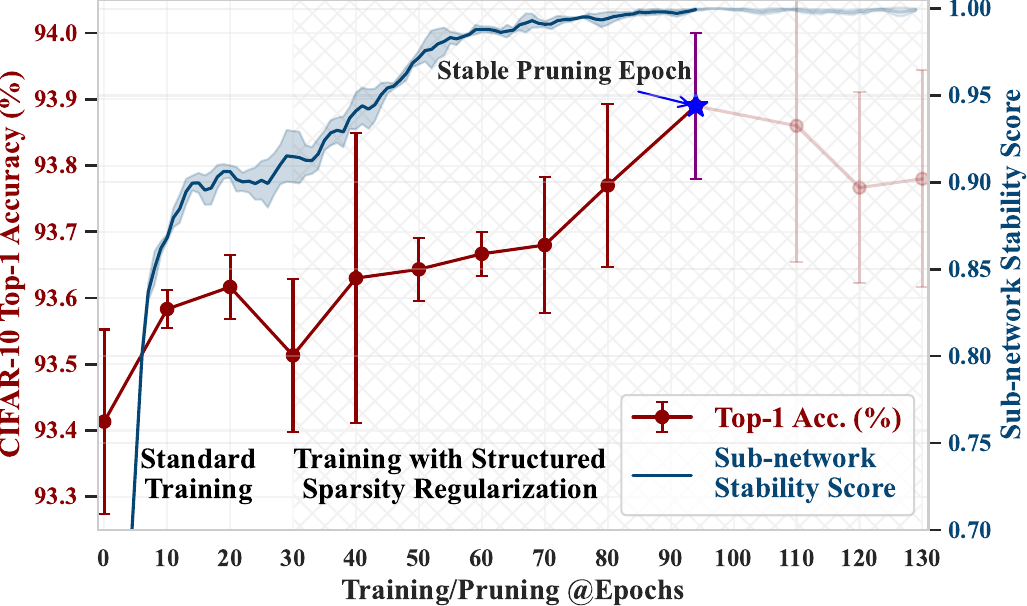}
\caption{Tracking the sub-network stability score and the corresponding pruned model's final accuracy over epochs for ResNet56 pruning on CIFAR-10 with a 50\% FLOPs reduction rate.
}
\label{fig:pss_accuracy_connection}
\end{figure}

\subsection{Ablation Studies}

\noindent \textbf{Correlation between Sub-network Stability Score and final Accuracy}.
\cref{fig:pss_accuracy_connection} illustrates the correlation between the stability score and the final accuracy of models pruned at various epochs until the training process reaches a stable pruning epoch. We measured the final accuracy of each temporarily pruned model obtained during the optimal sub-network search process guided by the pruning stability indicator. The \cref{fig:pss_accuracy_connection} clearly shows that as the stability score approaches its highest value, the best-performing pruned sub-network is also selected. The \cref{fig:pss_accuracy_connection} also shows that if we further delay the pruning process even after the stability score has peaked, the final accuracy of the pruned network begins to decline. This occurs because, once the pruning has stabilized, the structure of the pruned sub-network hardly changes, and there will be fewer and fewer training epochs left for fine-tuning the pruned network.

\noindent \textbf{Influence of Varying Sparsity Learning Start Epoch ($t_\text{sl-start}$)}. The start epoch for sparsity learning significantly influences the proposed method's overall training cost and performance. The algorithm behavior concerning training speed and performance for different values of $t_\text{sl-start}$ is illustrated in \cref{tab:t_sl_start_abal}. After a few initial training epochs, superior results were achieved by applying sparsity learning. Once the proposed stability indicator shows minimal variation across consecutive epochs, it indicates an optimal time to commence sparsity learning, as indicated by \cref{eqn:sl_start}. The \cref{tab:t_sl_start_abal} shows that delaying this process only leads to performance degradation along with increased training time.

\begin{table}[tb]
\caption{The influence on varying Sparsity Learning (SL) start epoch ($t_\text{sl-start}$) while pruning ResNet50 on the ImageNet. Given $t_\text{sl-start}$, $t^*$, i.e., stable pruning epoch, was 
identified using \cref{eqn:spe}.}
\label{tab:t_sl_start_abal}
\vspace{-5pt} 
\centering
\small 
\setlength{\tabcolsep}{6.5pt} 
\begin{tabular}{ccccc}
\toprule
\begin{tabular}[c]{@{}c@{}c@{}}SL Start \\ Epoch \\($t_\text{sl-start}$)\end{tabular} & 
\begin{tabular}[c]{@{}c@{}c@{}}SL End \\ Epoch \\ $(t^*$)\end{tabular} & 
\begin{tabular}[c]{@{}c@{}c@{}}Total SL \\ Epochs \\ ($t^* - t_\text{sl-start}) $\end{tabular} & 
\begin{tabular}[c]{@{}c@{}}Top-1 \\Acc. (\%)\end{tabular} & 
\begin{tabular}[c]{@{}c@{}}Training \\Speed Up \end{tabular} \\
\midrule
0 & 26 & 26 & 75.54 & 1.28$\times$ \\
15 & 31 & 16 & 75.90 & 1.28$\times$ \\
30 & 53 & 23 & 75.93 & 1.19$\times$ \\
45 & 69 & 24 & 75.81 & 1.13$\times$ \\
60 & 86 & 26 & 75.67 & 1.08$\times$ \\
17$^\dagger$ & 36 & 19 & \textbf{75.95} & \textbf{1.30}$\times$ \\
\bottomrule
\end{tabular}
\vspace{2pt}
\begin{minipage}{\linewidth}
\raggedright
\footnotesize
$^\dagger$ Automatically estimated using~\cref{eqn:sl_start}; all other \textit{SL Start Epoch} values were manually assigned.
\end{minipage}
\end{table}

\noindent \textbf{Pruning Pre-trained Models}. While OCSPruner is primarily designed for pruning from scratch, we also applied it to pre-trained models to reflect practical scenarios. Using the same training and pruning settings, we replaced randomly initialized models with pre-trained ones. As shown in~\cref{tab:pretraind_ocspruner}, OCSPruner achieves significant compression on VGG16 and ResNet56 (CIFAR-10) and ResNet50 (ImageNet), with minimal accuracy loss. This shows its effectiveness for randomly initialized and pre-trained models, with potential for further gains via hyperparameter tuning.

\begin{table}[tb]
\caption{Pruning pre-trained models with OCSPruner.}
\label{tab:pretraind_ocspruner}
\vspace{-5pt} 
\centering
\small 
\setlength{\tabcolsep}{1.5pt} 
\begin{tabular}{llccccc}
\toprule
Dataset & Model & \begin{tabular}[c]{@{}c@{}}FLOPs\\(\%)\end{tabular} & \begin{tabular}[c]{@{}c@{}}Params\\(\%) \end{tabular} & \begin{tabular}[c]{@{}c@{}}Base\\Top1\\Acc. (\%)\end{tabular} & \begin{tabular}[c]{@{}c@{}}Pruned\\Top1\\Acc. (\%)\end{tabular} & \begin{tabular}[c]{@{}c@{}}Top1\\Acc.\\$\downarrow$ (\%)\end{tabular} \\
\midrule
\multirow{2}{*}{CIFAR-10} & VGG16 & 21.21 & 13.68 & 94.07 & 93.63 & 0.44 \\
 & ResNet56 & 38.82 & 42.26 & 94.01 & 93.50 & 0.51 \\
\midrule
CIFAR-100 & VGG19 & 11.22 & 10.06 & 73.58 & 69.98 & 3.64 \\
\midrule
ImageNet & ResNet50 & 51.73 & 56.05 & 76.29 & 76.22 & 0.07 \\
\bottomrule
\end{tabular}
\end{table}


\mycomment{
\section{Conclusion}
\label{sec:conclusion}
In this work, we introduce an efficient and stable pruning framework named~\OursApproach{}.{Unlike traditional methods that rely on fully trained networks for pruning, our approach integrates pruning early in the training process, achieving competitive performance within a single training cycle.}~\OursApproach{} automatically identifies stable pruning architecture in the early training stage guided by the pruning stability indicator. An integral aspect of any pruning procedure lies in determining the optimal sub-network based on parameter saliency criteria. Consequently, our future research will focus on exploring better saliency criteria suitable for our pruning framework, surpassing the current reliance on simple norm-based criteria. Furthermore, we aim to evaluate and adapt our pruning methodology for diverse tasks such as object detection, segmentation, diffusion models, and large language models.

\noindent\textbf{Limitation.} 
While our method effectively uses regularization to eliminate redundant structures, it requires careful hyperparameter tuning tailored to each architecture and task. The algorithm is best suited for training schedules with gradual changes, such as cosine or linear decay, as they enable more stable pruning by avoiding abrupt shifts in learning rates. 
}

\section{Conclusion}
We presented OCSPruner, a novel one-cycle structured pruning framework that integrates training and pruning into a single efficient process. By introducing a stability-driven sub-network search guided by a pruning stability indicator, our method enables early and effective pruning. Through structured sparsity regularization and adaptive penalty scheduling, OCSPruner ensures both stable pruning and high model accuracy. Experiments across CIFAR and ImageNet benchmarks demonstrate that OCSPruner achieves state-of-the-art performance in terms of accuracy and training efficiency, outperforming several competitive methods. Moreover, it remains versatile and effective for both from-scratch and pre-trained models. Future work will investigate alternative saliency metrics and broader applications to modern architectures and other domains.

\noindent\textbf{Limitation.} 
Although our method removes redundant structures via regularization, it requires careful, task- and architecture-specific hyperparameter tuning. The algorithm is best suited for training schedules with gradual changes (e.g., cosine or linear decay), as they enable stable pruning by avoiding abrupt shifts in learning rates. 

\vspace{5pt}

\noindent\textbf{Acknowledgment.}
This work was supported by the National Research Foundation of Korea(NRF) grant funded by the Korea government(MSIT) (RS-2025-16071992), the convergence security core talent training business support program(IITP-2024 2024-RS-2024-00426853) supervised by the IITP(Institute of Information \& Communications Technology Planning \& Evaluation), and G-LAMP Program of the National Research Foundation of Korea (NRF) grant funded by the Ministry of Education (No. RS-2025-25441317). 
Additionally, this work was also supported by the Technology Innovation Program (20018906, Development of autonomous driving collaboration control platform for commercial and task assistance vehicles) funded by the Ministry of Trade, Industry and Energy (MOTIE, Korea).
{
    \small 
    \bibliographystyle{ieeetr}
    \bibliography{main}
}

\setcounter{page}{1}
\maketitlesupplementary

\section{Additional Ablation Studies}

\subsection{Group Saliency vs Conventional Saliency}

Conventionally, filter saliency estimation evaluates each convolutional layer in isolation when ranking filters based on their importance. In this approach, filters are assessed solely based on their individual contributions to the output of their respective layers, without accounting for interdependencies with other components in the network. In contrast, the group saliency used in our paper considers all parameters within a group, including those from interconnected convolutional layers, batch normalization layers, and fully connected layers.

Given the varied dimensions and scales of different group parameters, as described in \cref{sec:group_saliency_criteria}, we apply local normalization before computing the group saliency scores. This step ensures a fair comparison across heterogeneous parameter types and prevents components with inherently larger magnitudes from being overemphasized. As shown in \cref{tab:saliency_conv_vs_group}, pruning based on group saliency consistently outperforms conventional saliency estimation, leading to better accuracy-compression trade-offs.

\begin{table}[b]
\caption{Performance comparison on CIFAR-10 using OCSPruner with group vs. conventional saliency.}
\label{tab:saliency_conv_vs_group}
\vspace{-5pt}
\centering
\small
\resizebox{\columnwidth}{!}{%
\begin{tabular}{lccc|ccc}
\toprule
\multirow{2}{*}{Model} &
\multicolumn{3}{c|}{Group Saliency} &
\multicolumn{3}{c}{Conventional Saliency} \\ \cmidrule(lr){2-4} \cmidrule(lr){5-7}
& \begin{tabular}[c]{@{}c@{}}FLOPs\\(\%)\end{tabular} 
& \begin{tabular}[c]{@{}c@{}}Params\\(\%)\end{tabular} 
& \begin{tabular}[c]{@{}c@{}}Acc.\\(\%)\end{tabular} 
& \begin{tabular}[c]{@{}c@{}}FLOPs\\(\%)\end{tabular} 
& \begin{tabular}[c]{@{}c@{}}Params\\(\%)\end{tabular} 
& \begin{tabular}[c]{@{}c@{}}Acc.\\(\%)\end{tabular} \\
\midrule
Vgg16 & 19.99 & 7.18 & 93.87 & 20.00 & 3.75 & 92.94 \\
ResNet32 & 50.09 & 57.88 & 92.94 & 50.04 & 38.61 & 92.60 \\
ResNet56 & 46.88 & 47.66 & 93.85 & 47.03 & 43.93 & 93.48 \\
\bottomrule
\end{tabular}
}
\end{table}

\subsection{Analysis of Structured Sparsity Regularization}
\label{sec:sparsity}

\begin{figure}[t]
\centering
\includegraphics[width=1.0\linewidth]{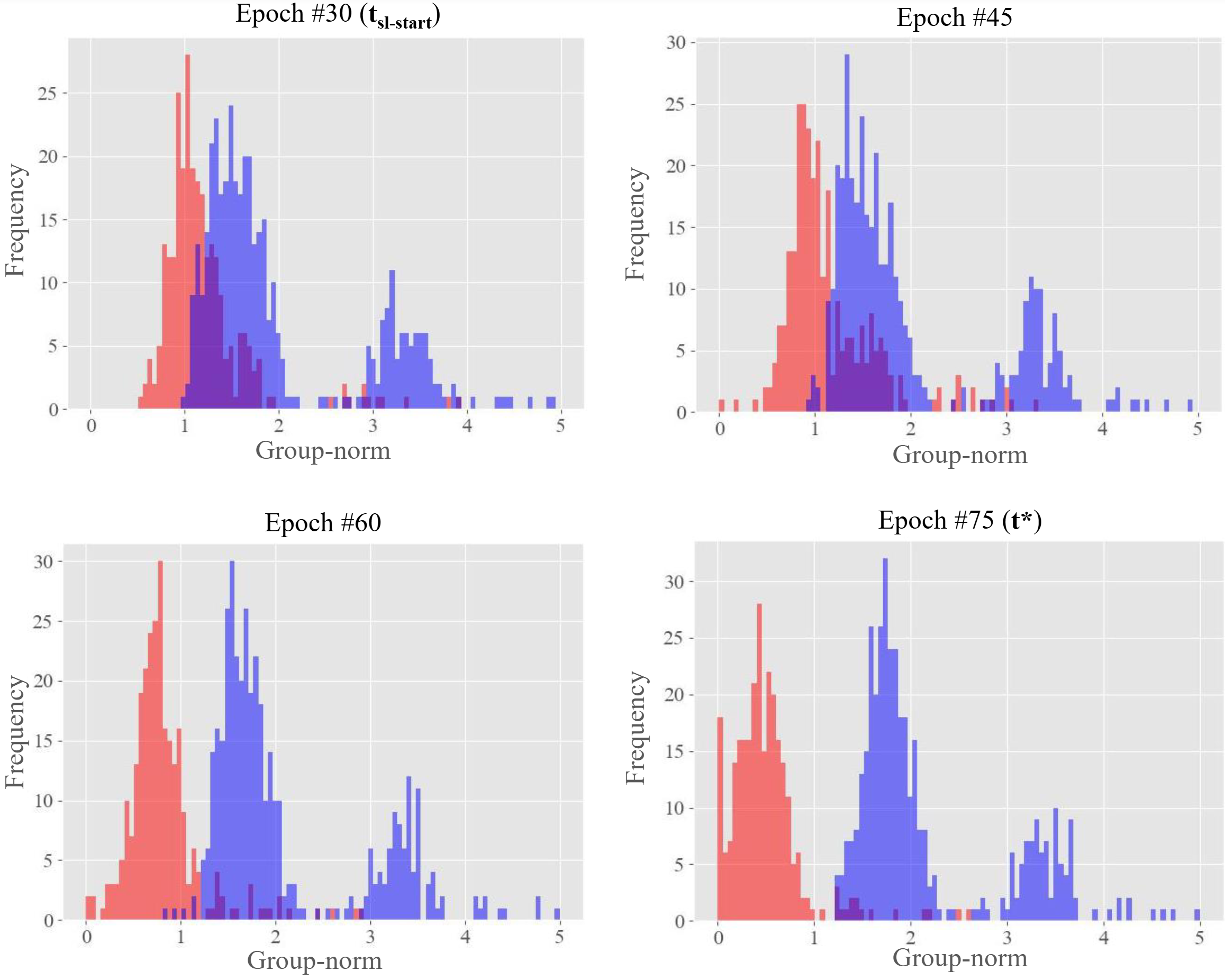}
\caption{Group-norm distribution of weight parameters while pruning ResNet32 model on CIFAR-10 dataset. Using structured sparsity learning (\cref{eqn:l2penalty} and~\cref{eqn:directmultiply}), the proposed method gradually drives grouped parameters (highlighted in red) toward zero.}
\label{fig:sl_progress}
\end{figure}

Regularization in the proposed pruning pipeline helps the stability indicator reach a stable epoch for final pruning. In particular, we utilize~\cref{eqn:l2penalty} and ~\cref{eqn:directmultiply} from~\cref{sec:sparsity_reg} of the main paper for this purpose. While~\cref{eqn:l2penalty} exhibit a slower sparsity learning,~\cref{eqn:directmultiply} is designed to speed up this process. \cref{tab:reg_abla} presents accuracy results for comparable FLOPs pruning rates, considering the application of regularization with~\cref{eqn:l2penalty} and~\cref{eqn:directmultiply} independently, and in combination. It is important to highlight that the parameter reduction rate is automatically determined based on the pruned architecture for a given FLOPs reduction rate. The accuracy values in the table indicate that better results are achieved when both equations~\cref{eqn:l2penalty} and~\cref{eqn:directmultiply} are used together for structured sparsity regularization.
\cref{fig:sl_progress} illustrates the progress of sparsity learning, starting from the sparsity learning initiation epoch ($t_\text{sl-start}$) up to the epoch of stable pruning ($t^*$).

\begin{table}[tb]
\caption{Ablation study on pruning accuracy using \cref{eqn:l2penalty} and \cref{eqn:directmultiply} independently and in combination for structured sparsity regularization.}
\label{tab:reg_abla}
\vspace{-5pt}
\centering
\small
\resizebox{\columnwidth}{!}{%
\begin{tabular}{lcccccc}
\toprule
\multirow{3}{*}{\begin{tabular}[c]{@{}l@{}}Model/\\Dataset\end{tabular}} &
\multicolumn{2}{c}{\multirow{2}{*}{\begin{tabular}[c]{@{}c@{}}Sparsity\\Learning\end{tabular}}} &
\multirow{3}{*}{\begin{tabular}[c]{@{}c@{}}FLOPs\\(\%)\end{tabular}} &
\multirow{3}{*}{\begin{tabular}[c]{@{}c@{}}Params\\(\%)\end{tabular}} &
\multirow{3}{*}{\begin{tabular}[c]{@{}c@{}}Top-1\\Acc.\\(\%)\end{tabular}} &
\multirow{3}{*}{\begin{tabular}[c]{@{}c@{}}Top-5\\Acc.\\(\%)\end{tabular}} \\
& & & & & & \\
\cmidrule{2-3}
& \cref{eqn:l2penalty} & \cref{eqn:directmultiply} & & & & \\
\midrule
\multirow{3}{*}{\shortstack{ResNet32/\\CIFAR-10}} 
& \checkmark & $\times$ & 44.69 & 38.59 & 92.09 & - \\
& $\times$ & \checkmark & 44.74 & 46.94 & 92.47 & - \\
& \checkmark & \checkmark & 44.98 & 48.28 & 92.74 & - \\
\midrule
\multirow{3}{*}{\shortstack{ResNet18/\\ImageNet}} 
& \checkmark & $\times$ & 55.09 & 55.13 & 67.89 & 88.00 \\
& $\times$ & \checkmark & 55.08 & 66.55 & 68.52  & 88.31 \\
& \checkmark & \checkmark & 55.04 & 67.90 & 68.66 & 88.61 \\
\bottomrule
\end{tabular}
}
\end{table}

\section{Training and Pruning Details}
\subsection{Hyper-parameter Configuration}

\cref{tab:hyperparameter} provides the training and pruning hyperparameters used for experiments on the CIFAR and ImageNet datasets. For the CIFAR dataset, the same parameters are utilized regardless of the model used. However, for the ImageNet dataset, slightly different training settings are employed for the ResNet50 and MobileNetV2 models.

\begin{table}[tb]
\caption{Training and pruning hyper-parameters for OCSPruner.}
\label{tab:hyperparameter}
\vspace{-5pt} 
\centering
\small 
\setlength{\tabcolsep}{-1pt}
\begin{tabular}{llcc}
\toprule
\multirow{2}{*}{{Task}} & \multirow{2}{*}{{\hspace{1.0em}Parameters}} & \multicolumn{2}{c}{{\hspace{-1.0em}Dataset}} \\
\cmidrule(lr){3-4}
 & & \hspace{-2.5em}CIFAR & ImageNet \\
\midrule
\multirow{11}{*}{\rotatebox[origin=c]{90}{Training}} & \hspace{1.0em}batch size & \hspace{-2.5em}128 & 256 \\
& \hspace{1.0em}epochs & \hspace{-2.5em}300 & \begin{tabular}[r]{@{}r@{}}ResNet50: 130\\MobileNetV2: 150\end{tabular} \\
& \hspace{1.0em}optimizer & \hspace{-2.5em}SGD & SGD \\
& \hspace{1.0em}momentum & \hspace{-2.5em}0.9 & 0.9 \\
& \hspace{1.0em}learning rate & \hspace{-2.5em}0.1 & \begin{tabular}[r]{@{}r@{}}ResNet50: 0.1\\MobileNetV2: 0.05\end{tabular} \\
& \hspace{1.0em}\begin{tabular}[l]{@{}l@{}}learning rate\\decay  \end{tabular}& \hspace{-2.0em}\begin{tabular}[c]{@{}c@{}}MultiStepLR\\(90, 180, 240, 270; 0.2)\end{tabular} & \hspace{-1.0em}Cosine \\
& \hspace{1.0em}weight decay & \hspace{-2.5em}5e-4 & 4e-5 \\
\midrule
\multirow{6}{*}{\rotatebox[origin=c]{90}{Pruning}} & \hspace{1.0em}threshold for $t_\text{sl-start}$ ($\tau$) & \hspace{-2.5em}- & \hspace{-1.0em}1e-4 \\
& \hspace{1.0em}$t_\text{sl-start}$ & \hspace{-2.5em}30 & \hspace{-1.0em}auto (Eq. (4)) \\
& \hspace{1.0em}threshold for $t^*$ ($\epsilon$) & \hspace{-2.5em}1e-3 & \hspace{-1.0em}1e-3 \\
& \hspace{1.0em}\begin{tabular}[l]{@{}l@{}}reg. penalty\\initial value ($\lambda_0$) \end{tabular} & \hspace{-2.5em}1e-4 & \hspace{-1.0em}1e-4 \\
& \hspace{1.0em}\begin{tabular}[l]{@{}l@{}}reg. penalty\\increment value ($\sigma$) \end{tabular} & \hspace{-2.5em}1e-4 & \hspace{-1.0em}1e-4 \\
& \hspace{1.0em}\begin{tabular}[l]{@{}l@{}}$\lambda$ increment\\epoch interval ($\Delta t$) \end{tabular} & \hspace{-2.5em}1 & \hspace{-1.0em}2 \\
\bottomrule
\end{tabular}
\end{table}

\subsection{Computational Cost Analysis}

We performed algorithm complexity analysis on a PC with a 64-bit Ubuntu 20.04.6 LTS operating system, driven by an AMD Ryzen 9 7950X 16-Core Processor, 62 GB RAM, and two NVIDIA GeForce RTX 4090 GPUs. Notably, all experiments utilized a single GPU except for ResNet50 pruning on ImageNet, where both GPUs were used. \cref{tab:complexity_pruning} presents OCSPruner's time complexity, highlighting its efficiency relative to baseline training cost, particularly for large-scale datasets where training time is crucial. While the OCSPruner cost for CIFAR-10/100 datasets slightly surpasses the base network training cost, it is crucial to acknowledge that network slimming offers negligible advantages in small-scale dataset training time, given that the pruning process cost is already factored into the overall calculation.

\begin{table}[tb]
\caption{The comparison of OCSPruner overall cost versus baseline model training cost.}
\label{tab:complexity_pruning}
\vspace{-5pt} 
\centering
\small 
\resizebox{\columnwidth}{!}{%
\begin{tabular}{llcc}
\toprule
Dataset & Model & \begin{tabular}[c]{@{}c@{}}Baseline Model\\Training Time\end{tabular} & \begin{tabular}[c]{@{}c@{}}OCSPruner\\Total Time \end{tabular} \\
\midrule
\multirow{2}{*}{CIFAR-10} & VGG16 & 0 h 25 min & 0 h 27 min \\
 & ResNet56 & 0 h 25 min & 0 h 26 min \\
\midrule
CIFAR-100 & VGG19 & 0 h 26 min & 0 h 30 min \\
\midrule
\multirow{2}{*}{ImageNet} & \multirow{1}{*}{ResNet50} & \multirow{1}{*}{37 h 55 min} & 27 h 6 min \\
 & \multirow{1}{*}{MobileNetV2} & \multirow{1}{*}{38 h 20 min} & 31 h 57 min \\
\bottomrule
\end{tabular}
}
\end{table}

\section{Inference Efficiency on Edge Devices}
\label{sec:latency}
The primary aim of structurally pruning baseline models is to make them suitable for deployment on low-powered edge devices, resulting in reduced model size and inference time. To validate this, we conducted experiments using the NVIDIA Jetson Xavier NX board to evaluate the inference time of our pruned models on such a low-powered device.
In our evaluations, as summarized in~\cref{tab:latency}, we use a batch size of 128 for the CIFAR-10/100 datasets and a batch size of 16 for the ImageNet dataset. We conducted 200 inference runs in each test case and averaged the latency time. For example, pruning the ResNet50 model FLOPs by 67\% on the ImageNet dataset results in a practical speedup of 2.45$\times$ compared to the baseline model's latency. However, the theoretical speedup is expected to be 3.03$\times$. Similarly, in practice, we can see that an 89\% FLOPs reduction in the VGG19 model on the CIFAR-100 dataset leads to a 4.22$\times$ speedup in inference latency, whereas the theoretical speedup is around 8.5$\times$.

\begin{table*}[tb]
\caption{Latency analysis comparing baseline and pruned models on various datasets using the NVIDIA Jetson board. The percentage value in the subscript denotes the FLOPs pruning rate.}
\label{tab:latency}
\centering
\small 
\begin{tabular}{lllcccccc}
\toprule
\multirow{2}{*}{\centering Dataset} &  
\multirow{2}{*}{\centering Model} &
\multirow{2}{*}{\centering Method} &
\multirow{2}{*}{\centering MACs (G)} & 
\multirow{2}{*}{\centering Params (M)} &
\multicolumn{3}{c}{\centering NVIDIA Jetson Xavier NX} \\ \cmidrule(lr){6-8}
&&&&& \centering Batch Size & \centering Latency (ms) & \centering Speed Up & \\ \midrule
\multirow{7}{*}{CIFAR-10} & \multirow{2}{*}{ResNet56} & Baseline & 0.13 & 0.86 & 128 & 97.2 & 1.00$\times$\\
& & $\text{OCSPruner}_{61\%}$ & 0.05 & 0.32 & 128 & 63.6 & 1.54$\times$\\
\cmidrule{2-9}
 & \multirow{2}{*}{ResNet110} & Baseline & 0.26 & 1.73 & 128 & 193.0 & 1.00$\times$ \\
& & $\text{OCSPruner}_{67\%}$ & 0.08 & 0.51 & 128 & 110.5 & 1.75$\times$ \\
\cmidrule{2-9}
 & \multirow{2}{*}{VGG16} & Baseline & 0.40 & 14.73 & 128 & 90.8 & 1.00$\times$ \\
& & $\text{OCSPruner}_{80\%}$ & 0.08 & 1.0 & 128 & 31.2 & 2.90$\times$ \\
\midrule
\multirow{2}{*}{CIFAR-100} & \multirow{2}{*}{VGG19} & Baseline & 0.51 & 20.09 & 128 & 114.2 & 1.00$\times$ \\
& & $\text{OCSPruner}_{89\%}$ & 0.06 & 1.15 & 128 & 27.3 & 4.22$\times$ \\
\midrule
\multirow{5}{*}{ImageNet} & \multirow{5}{*}{ResNet50} & Baseline & 4.12 & 25.56 & 16 & 256.9 & 1.00$\times$ \\
& & $\text{OCSPruner}_{43\%}$ & 2.35 & 16.47 & 16 & 164.4 & 1.56$\times$ \\
& & $\text{OCSPruner}_{48\%}$ & 2.13 & 15.89 & 16 & 154.5 & 1.66$\times$ \\
& & $\text{OCSPruner}_{57\%}$ & 1.78 & 13.25 & 16 & 133.1 & 1.93$\times$ \\
& & $\text{OCSPruner}_{67\%}$ & 1.36 & 10.41 & 16 & 105.3 & 2.45$\times$ \\
\bottomrule
\end{tabular}
\end{table*}

\section{Applying OCSPruner to ViT \& DeiT Models}

The OCSPruner is also used to prune Vision Transformer (ViT) [1]  architecture, which has gained significant attention in computer vision. We integrate OCSPruner into the widely-used DeiT training framework for image classification by applying data augmentation techniques similar to DeiT [2]. The training uses the AdamW optimizer with a cosine learning rate scheduler over 300 epochs, a batch size of 1024, a weight decay of 0.05, and an initial learning rate of 5e-4. For ViT, a warm-up period of 10 epochs with a warm-up learning rate of 0.001$\times$lr is used, while for DeiT, pruning is followed by training with a 5-epoch warm-up at a warm-up learning rate of 0.1$\times$lr. Remarkably, our method achieved a 30-40\% reduction in FLOPs while training from scratch on ImageNet. Despite this substantial reduction in FLOPs, the pruned models retained performance comparable to the original counterparts, and in the case of DeiT, the pruned model even showed improved performance.

\begin{table}[tb]
\caption{Pruning ViT and DeiT models on ImageNet-1k.}
\label{tab:vit-deit-pruning-imagenet}
\vspace{-2pt} 
\centering
\small 
\resizebox{\columnwidth}{!}{%
\begin{tabular}{llccccc}
\toprule
Model & Method & \begin{tabular}[c]{@{}c@{}}Params \\(M)\end{tabular} & \begin{tabular}[c]{@{}c@{}}FLOPs \\(G)\end{tabular} & \begin{tabular}[c]{@{}c@{}}Top-1 \\ACC (\%)\end{tabular} & \begin{tabular}[c]{@{}c@{}}Top-5 \\ACC (\%)\end{tabular} \\ 
\midrule
\multirow{2}{*}{ViT-Small} & Baseline & 22.05 & 4.61 & 78.84 & 95.33 \\
& OCSPruner & 12.18 & 2.72 & 77.59 & 93.31 \\ 
\midrule
\multirow{2}{*}{DeiT-Tiny} & Baseline & 5.72 & 1.26 & 72.20 & 91.10 \\
& OCSPruner & 3.82 & 0.88 & 72.51 & 91.15 \\
\bottomrule
\end{tabular}
}
\end{table}

\section{Extended Applications: Object Detection}
\label{sec:objdet}

In this experiment, we applied a pruned ResNet50 model with OCSPruner as a backbone network for object detection. We use Single Shot multibox Detector (SSD) [3] as our detection network with an input size resolution of 300 × 300. The COCO object detection dataset [4] is used for training and evaluation which contains images of 80 object categories with 2.5 million labeled instances in 328k images. The pruned ResNet50 backbone model trained with OCSPruner on the ImageNet dataset is attached with SSD head and finetuned on the COCO object detection dataset. The model is trained for 130 epochs with a batch size of 32, a learning rate of 0.0026, and a weight decay of 0.0005, with the learning rate decayed at epochs 86 and 108, utilizing two NVIDIA GeForce RTX 4090 GPUs. 

The performance of SSD300 with a pruned ResNet50 backbone is detailed in \cref{tab:object_detection}. In our experimental setup using the default ResNet50 backbone, we achieved an average precision (AP) of 26.90\%. Transitioning to a pruned ResNet50 backbone resulted in a significant reduction in FLOPs compared to parameter reduction rates. By simplifying the backbone architecture, we achieved a notable overall complexity reduction, albeit requiring channel pruning in the SSD head for substantial parameter reduction. It's important to note that our experiments focused solely on varying pruning rates for the backbone while keeping the SSD head unchanged. Remarkably, we observed only a 1.12\% drop in AP with a 25.42\% reduction in FLOPs. Furthermore, an approximately 3\% drop in AP corresponded to a 56\% reduction in overall FLOPs.

\begin{table}[t]
\caption{Object Detection results using SSD300 with a pruned ResNet50 backbone on the COCO dataset.}
\label{tab:object_detection}
\vspace{-8pt} 
\centering
\small 
\resizebox{\columnwidth}{!}{%
\begin{tabular}{ccccc}
\toprule
\begin{tabular}[c]{@{}c@{}}Backbone\\(\%)\end{tabular} & 
\begin{tabular}[c]{@{}c@{}}FLOPs\\(G) (\%)\end{tabular} & 
\begin{tabular}[c]{@{}c@{}}Params\\(M) (\%)\end{tabular} & 
\begin{tabular}[c]{@{}c@{}}Average \\Precision (AP)\end{tabular} & 
\begin{tabular}[c]{@{}c@{}}Average \\Recall (AR)\end{tabular} \\ 
\midrule
ResNet50 & 100 & 100 & 26.90 & 38.10 \\
\midrule
\multirow{3}{*}{\begin{tabular}[c]{@{}c@{}}Pruned\\ResNet50\end{tabular}} & 74.58 & 86.63 & 25.78 & 37.01 \\
 & 65.67 & 81.52 & 25.30 & 36.50 \\
 & 44.06 & 68.11 & 23.82 & 35.44 \\
\bottomrule
\end{tabular}
}
\end{table}

\section{Visualization of Pruned Network Structures}
\label{sec:netviz}
\label{sec:addivis}
\begin{figure*}[tb]
\centering
\includegraphics[width=0.95\linewidth]{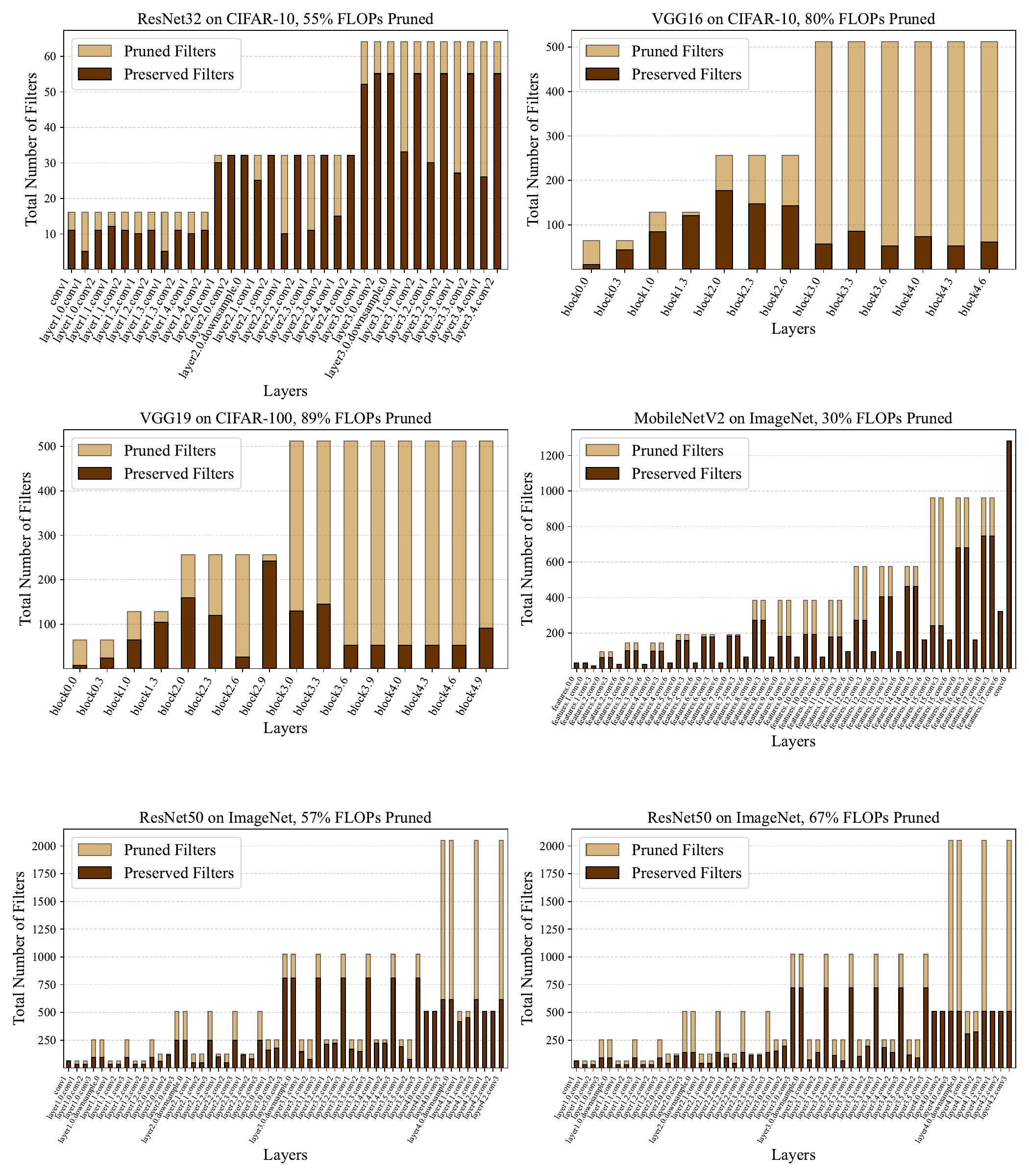}
\caption{Visualization of pruned models displaying pruned and retained filters across each layer's total number of filters.}
\label{fig:vis_pruning_models}
\end{figure*}

The visualization in~\cref{fig:vis_pruning_models} shows the result of pruning across different models, pruning rates, and datasets with OCSPruner. Baseline models often exhibit unnecessary complexity that is typically not required. As illustrated in~\cref{fig:vis_pruning_models}, when pruning VGG16/19, numerous filters from the later parts of the network are eliminated, whereas when pruning ResNet models, filters are consistently removed from all parts of the network. Yet, the pruned networks continue to perform well. This observation suggests that layers with higher pruning rates capture minimal or redundant information, rendering them suitable for pruning.

\end{document}